\documentclass[runningheads]{llncs}
\usepackage{graphicx}
\usepackage{multirow}
\usepackage{amsmath}
\usepackage{xcolor}
\begin{document}
\title{Interactive Deep Refinement Network for Medical Image Segmentation}
%
%
\author{Titinunt Kitrungrotsakul\inst{1} \and
Iwamoto Yutaro\inst{2} \and
Lanfen Lin \inst{3} \and
Ruofeng Tong \inst{1,3} \and
Jingsong Li\inst{1,4} \and
Yen-Wei Chen,\inst{1,2} }
\authorrunning{T. Kitrungrotsakul, I. Yutaro, L. Lin,  et al.}
%

\institute{Healthcare Data Science, Zhejiang Lab, Hangzhou, China \email{titinuntkitrungrotsakul@zhejianglab.com} \and
Graduate School of Information Science and Engineering, Ritsumeikan University, Shiga, Japan \\ \and
College of Computer Science and Technology, Zhejiang University, Hangzhou, China\and
College of Biomedical Engineering \& Instrument Science, Zhejiang University, Hangzhou, China}
\maketitle              
\begin{abstract}
Deep learning techniques have successfully been employed in numerous computer vision tasks including image segmentation. The techniques have also been applied to medical image segmentation, one of the most critical tasks in computer-aided diagnosis. Compared with natural images, the medical image is a gray-scale image with low-contrast (even with some invisible parts). Because some organs have similar intensity and texture with neighboring organs, there is usually a need to refine automatic segmentation results. In this paper, we propose an interactive deep refinement framework to improve the traditional semantic segmentation networks such as U-Net and fully convolutional network. In the proposed framework, we added a refinement network to traditional segmentation network to refine the segmentation results.Experimental results with public dataset revealed that the proposed method could achieve higher accuracy than other state-of-the-art methods.

\end{abstract}
\section{INTRODUCTION}

Organ segmentation is one of the most crucial tasks in computer aided diagnosis (CAD) system. A wide range of computer vision techniques has been developed for medical image segmentation. These methods can be summarized as: region-based approaches \cite{Adams1994,Pan2004}, active contours \cite{Zhou2016,Foruzan2013}, graph-cut-based or random walk-based interactive methods \cite{Boykov2006,Grady2005,Yuan2018}, and anatomic models-based methods \cite{Okada2008,Dong2015}.

Recently, deep learning-based semantic segmentation such as the fully convolutional network (FCN) has been attracting considerable interest. U-Net et al. FCN \cite{Long2015} replaces the fully connected layers in the traditional convolutional neural network with convolutional layers to classify the image at the pixel level. U-Net \cite{Ronneberger2015,Abraham2018} consists of a contracting path that captures context and a symmetric expanding path that enables precise localization. In 2015, Google proposed DeepLab \cite{Chen2014}, and subsequently introduced DeepLab v2 \cite{Chen2018}, v3 \cite{Chen2017}, and v3+ \cite{Chen20182}; which included improvements such as atrous convolution, atrous spatial pyramid pooling. Deep learning-based segmentation methods achieve state-of-the-art performance in the computer vision domain. These models are also widely applied to medical image segmentation and achieved impressive performance \cite{Roth2017,He2019,Liu2018,Larsson2018,Tang2013}.

Compared with natural images, the medical image is a gray-scale image with low-contrast (even with some invisible parts). Because some organs have similar intensity and texture with neighboring organs, there is usually a need to refine automatic segmentation results. 

Graph Cuts\cite{Boykov2006} method is an interactive segmentation method and also can be used as refinement method. The problems of the graph cut occur when applied it to medical data. The graph cut method assigns each voxel as a node in the graph that will lead to large memory usage and high computational time for optimization.

Conditional random fields (CRFs) \cite{Chen2014} often applied in pattern recognition and machine learning. CRF is a discriminative undirected probabilistic graphical model that is used when the class labels for different input are not independent. This approach is widely used to refine segmentation results by deep learning based semantic segmentation as an independent post-processing technique.

Generative Adversarial Network (GAN) is an unsupervised deep learning network proposed by I. Goodfellow \cite{Goodfellow2014}. In GAN, the Generator network generates data that is similar to the training data. While discrimination is to identify the data from training data and the data from the Generator. The objective of GAN in segmentation model is used to refine the weight of segmentation network. Both CRF and GAN are used to refine segmentation network, however, CRF uses an independent indirect graph to refine the segmentation result. While GAN used as a weight refinement in the training phase and ignore segmentation result on test phase.

\begin{figure}[!tb]
  \centering
  \includegraphics[width=3in]{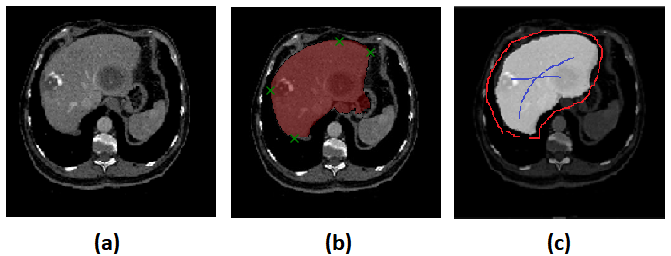}
  \caption {Seed points comparison between between DEXTR and other interactive methods, a) is a input image, b) is seed points (green) for DEXTR, c) seed points of other interactive methods.}
  \label{Fig:seed_compare}
\end{figure}

Deep Extreme Cut (DEXTR) \cite{DEXTR} is the recent deep learning method that uses the concept of interactive for image segmentation. The method requires only four extreme points in an object (left-most, right-most, top, bottom pixels) as extra input information. The method is different from CRFs and GANs, which DEXTR adding an extra channel to the input image of a convolutional neural network, which contains Gaussian centered in each of extreme points. The disadvantages of this network are users do not know the area of difficulty until they try, and users need to put extreme points cautiously, one pixel missing may curse a totally different segmentation result. The sample of seed points given in DEXTR is shown in Figure \ref{Fig:seed_compare}

In this paper, we propose an interactive deep refinement framework to improve the traditional semantic segmentation networks, such as U-Net and FCN. In the proposed framework, we added a deep refinement network to the traditional segmentation network to refine the segmentation results. The main advantages and contributions of the proposed method are as follows.

\begin{enumerate}
  \item Compared with CRF, which is widely used to refine segmentation results as an independent post-processing technique, the proposed refinement network is connected and trained together with the main segmentation network. Furthermore, the proposed refinement network is specific to and targeted at medical images.
  \item Compared with GAN, utilizing a discriminator to refine the segmentation network (generator), the trained refinement network is used directly to refine the segmentation results of a test image based on its seed points.
  \item We propose an automatic seed points generation method, allowing for end-to-end training without seed points provided by the user.
  \item Besides, the proposed interactive refinement network differs from other deep learning-based interactive segmentation methods such as Deep Extreme Cut (DEXTR) \cite{DEXTR}, in which four extreme points are extracted for each training samples to train the segmentation network. Generally interactive methods such as DEXTR or graph cut do not provide the difficult segmented region to the user moreover, they require more interactive to finish segmentation task. Motivated by the cons of interactive segmentation methods, we also provide in this paper, the region of the difficult segmented region that may require seed points from users.
\end{enumerate}



\section{Refinement segmentation}
\label{sec:method}

\subsection{Proposed Architecture}

\begin{figure*} [tb]
   \begin{center}
       \begin{tabular}{c} 

       \includegraphics[width=3.5in]{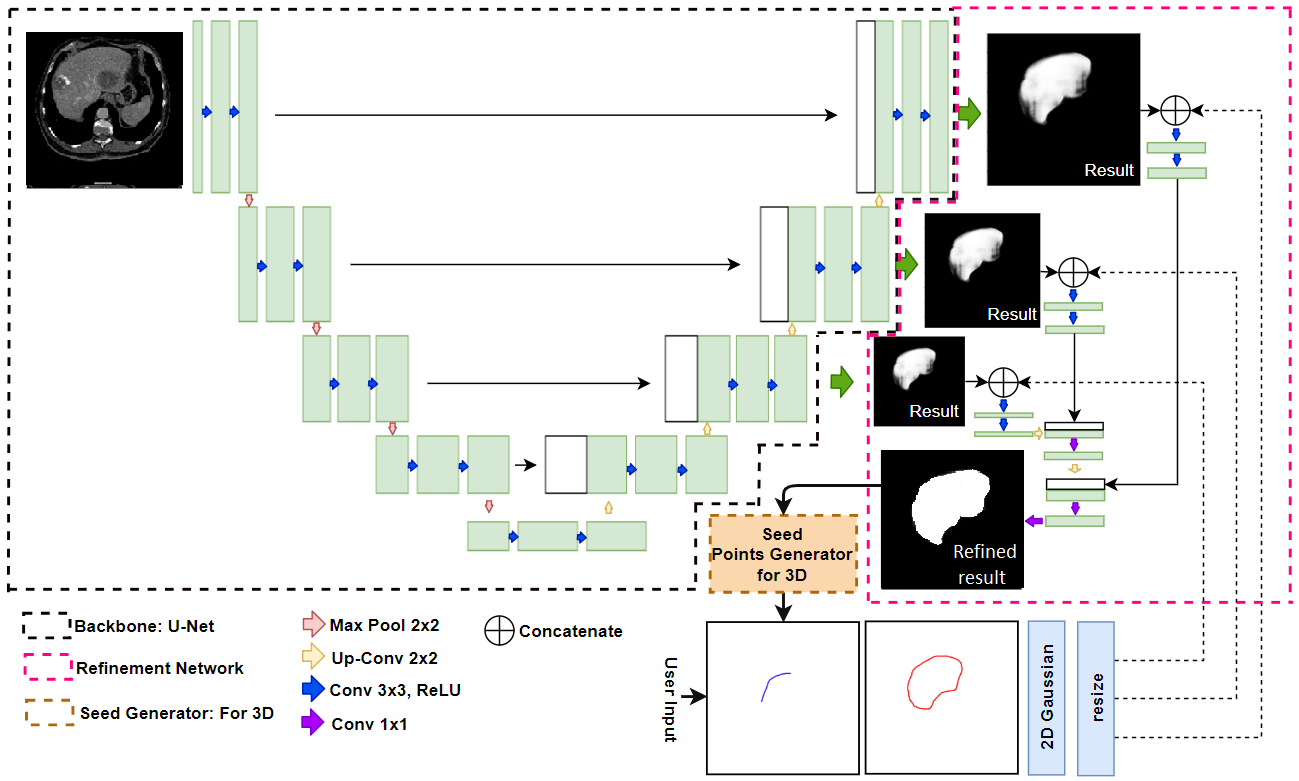}
       \end{tabular}
   \end{center}
   \caption[example]
   {\label{fig:network}
    The Architecture of our network: Input image is proceeded by the backbone network to produce
  the initial segmentation $p$ in the box indicated by black color. Both the initial segmentation $p$ and
  seed points are used to generate refined segmentation mask $\hat{p}$ by the refinement network in another
  box indicated by red color. The refined segmentation result $\hat{p}$ is used to generate seed points for
  neighbor slices in the case of 3D volume.}
\end{figure*}

Figure\ref{fig:network} shows the architecture of the proposed RefineNet. The network consists of two parts. The first part of the RefineNet is the backbone segmentation network which is used to extract initial segmentation. The second part is the refinement network which uses initial segmentation result of the first network and seed point information to generate refinement result.

The main objective of the proposed network is to refine the segmentation results of a test image based on its seed points. Unlike CRF, graph cut, and GANs, the proposed refinement network is connected and trained with the first part of the segmentation network.

Furthermore, we utilize the U-Net as our initial segmentation backbone, as its performance in medical image segmentation is better than that of other deep learning segmentation methods. Additionally, we use the concept of a pyramid network \cite{PSP} by adding the 1$\times$1 convolution to generate initial segmentation results before the last three deconvolutional layers which the size of initial segmentation results of last three layers are 1$\times$, 0.5$\times$, and 0.25$\times$ of original input image size. Each initial segmentation result concatenated with seed points images and merged up each scale together using concatenation method and deconvolutional (Up-Conv.). So the size of final segmentation result of second network is same as the size of input image. 


The output of the second network is the segmentation result that refined using seed points images as refinement guidelines. To use seed points as image guideline in the refinement network, we transform seed point marks using 2D Gaussian. The output of the network is a probability map for a whole image. Each pixel of the probability map contains the probability of foreground and background value.

\subsection{Seed Point Generation}

The seed points in our research are obtained from user input and auto-generated values. Because of the unpredictability of the initial segmentation from the network, manually provided seed points from users cannot be used in each epoch during the training phase. However, they can be used in the testing phase.

\textbf{Training: }To avoid random giving of seed points during the training by the user, which is due to the unknowing of the initial segmentation result from the first network, we provide a solution that does not require seed points during the training phase.

To generate the foreground/background points from the initial segmentation, we use both the ground truth ($f_y$) and initial segmentation result ($f_x$) generated from the first network. The subtraction of both images generates the subtraction mask ($\delta f$); indicating that each pixel is either $-1$, 0, or $+1$. The $+1$ mask is referred to as the over-segmentation mask, while the $-1$ mask is regarded as the under-segmentation mask.

\begin{equation}
    \delta f = f_x - f_y
\begin{cases}
    \delta f>0,  & \text{background point} \\
    \delta f<0,  & \text{foreground point}
\end{cases}
\end{equation}

\begin{figure}[tb]
  \centering
  \includegraphics[width=3.5in]{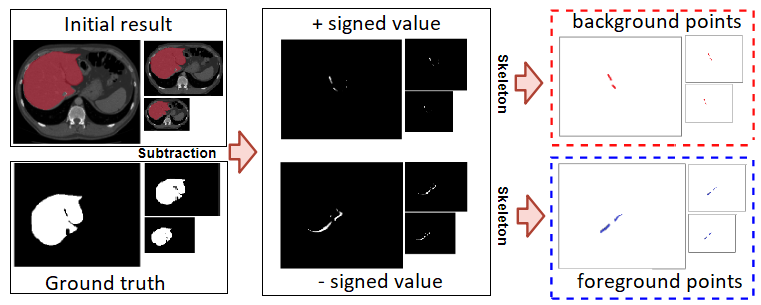}
  \caption {Seed points generation flow. Red boxes are background seed points generation. Blue boxes are foreground seed points generation.}
  \label{Fig:seedpoint}
\end{figure}
To generate seed points from those masks, we apply a skeleton to those masks and use the skeleton line as a foreground seed points (under-segmentation mask), and background seed points (over-segmentation mask) as shown in Figure\ref{Fig:seedpoint}.

\textbf{Testing: }The main disadvantage of generally interactive segmentation methods are users do not know the region of difficulty, which is mostly under- or over- segmentation until the initial segmentation result is obtained. In our method, the initial segmentation is generated from the first network. With the initial segmentation, the user can observe the segmented, over-segmentation, and under-segmentation region and uses them as input guideline. Unlike seed points in the training phase that auto-generated seed points require ground truth image, manual seed points are inputted from the user.

\section{Experiments and results}
\label{sec:EXPERIMENTS}

\subsection{Implementation Detail}
To develop our network, we use Keras and TensorFlow as our development libraries. The hyper-parameters of the network were tuned using cross-validation with a validation set \cite{Bengio2013}. The chosen hyper-parameters were those with which the model performed best with the validation set. All computations were performed using an NVIDIA GeForce GTX TITAN X GPU with 12 GB memory.


\subsection{Evaluation Methods}
The leave-one-out method was used in our experiments. We selected one CT volume as a test image, and other CT volumes were used for training. For each method, all data are performed to collect experimental results with a different training set. The overall performance of each method was evaluated using the test set. The overlap index (Dice coefficient), Sensitivity (SEN), and Positive Predictive Value (PPV) were used as evaluation measures.






The proposed method was evaluated using the IRCAD datasets, which are maintained by a French research institute \cite{IRCAD}.

\subsection{Liver segmentation benchmark}
In this section, we evaluate our network and benchmark on 2D images because most of the existing segmentation methods are done their work on 2D images. We slicing the 3D volume in the horizontal axis to obtain 2D images. With this method, we can obtain approximately 120 slices from each volume. All benchmark methods in this experiment perform segmentation on the 2D slice and form the 2D segmentation result in 3D segmentation volume.  The benchmark methods in this experiment are divided into 2 groups. First group is automatic segmentation methods which are Li et al. \cite{Li2013}, Cascaded U-Net \cite{Christ2016}, Cascaded U-Net with 3D CRF\cite{Christ2016} FCN\cite{Long2015}, and U-Net\cite{UNET}. Another group is interactive segmentation methods which are graph cut\cite{Boykov2006}, DEXTR\cite{DEXTR}, and our proposed method.

We retrain all deep learning methods using IRCAD dataset with the leave-one-out technique. For graph cut \cite{Boykov2006}, we use the boundary of dilation method on ground truth as background seed and apply skeletonization on ground truth for foreground seed. For DEXTR \cite{DEXTR} network, we use the top, bottom, leftmost, and rightmost points as the guideline points.

%

\begin{table*}[!bt]
\caption{ A comparison of dice values of the segmentation result on 3D volume.}

\begin{center}
\begin{tabular}{| c| c | c c c|}
  \hline	
        \quad Methods \quad&\quad Approach \quad&\quad DICE \quad&\quad SEN \quad&\quad PPV \quad\\\hline		
        Li et al.\cite{Li2013}  & Automatic & 0.945  & - & - \\
        FCN\cite{Long2015}  & Automatic  & 0.709  & 0.665& 0.743  \\
        U-Net\cite{UNET}  & Automatic  & 0.729  & 0.709 & 0.742 \\
        Cascaded U-Net\cite{Christ2016}  & Automatic  & 0.931  & - & - \\
        Cascaded U-Net + 3D CRF\cite{Christ2016}  & Automatic  & 0.943 & - & - \\
        Graph Cut\cite{Boykov2006}  & Interactive & 0.936  & 0.885 & 0.911  \\
        DEXTR\cite{DEXTR}  & Interactive & 0.907  & 0.917 & 0.853  \\       \hline
        RefineNet(FCN) & Interactive & 0.941 & 0.927 & 0.949 \\
        RefineNet(U-Net) & Interactive & {0.960} & 0.955 & 0.964\\
  \hline
\end{tabular}
\end{center}
\label{table:resnet_state}
\end{table*}

The experimental results are shown in Table \ref{table:resnet_state}. Graph cut and DEXTR methods use an interactive technique and can obtain high accuracies. However, DEXTR can obtain only 0.907 because the method is prepared for non-medical imaging and met the intensity problem, while the graph cut method is well prepared for medical imaging and obtain 0.936 of dice. In the automatic approach, Li et al., Cascaded U-Net and cascaded U-Net with 3D CRF methods can get very high accuracies which are 0.945, 0.931 and 0.943. Those methods designed for medical liver segmentation with well prepare to preprocess and post-processing, while U-Net and FCN cannot get good results same as other methods which are 0.729 and 0.709.

Based on those results in Table \ref{table:resnet_state}, FCN and U-Net are lowest accuracies than other methods. However, both method not include preprocessing and posprocessing in their process which can significantly improve. In this experiment, we applied our refinement method to refine FCN and U-Net results. The results of proposed refinement method on FCN and U-Net are significantly improve which are 0.941 and 0.960. 

In order to evaluate proposed network as a post-processing method, we compare proposed network with other post-processing methods which are CRF, GAN, and graph cut on the same segmentation backbone network.

\subsection{Semi-automatic slice by slice segmentation}
In this section, we evaluate our proposed method and other methods using semi-automatic segmentation. The input of all interactive segmentation methods used in this experiment is the one perfect liver segmentation slice ($Seg_i$) that slice located on the middle or largest liver slice in the volume. Where $i$ indicate the slice index in the volume. We use liver segmentation slice ($Seg_i$) to generate the seed points for interactive segmentation methods.

\begin{table}[!t]
\caption{ A comparison on slice by slice liver segmentation.}

\begin{center}
\begin{tabular}{| c|  c  c c|}
  \hline	
        \quad Methods \quad&\quad DICE \quad&\quad SEN \quad&\quad PPV \quad\\\hline		
        Graph Cut\cite{Boykov2006}  & 0.890 & 0.932 & 0.855  \\
        Graph Cut w shape\cite{Pipe_ICIP}  & 0.926 & 0.908 & 0.937  \\
        DEXTR\cite{DEXTR}  & 0.582 & 0.825 & 0.450  \\       \hline
        RefineNet(U-Net) & \textbf{0.937} & 0.941 & 0.918\\
  \hline
\end{tabular}
\end{center}
\label{table:slice_by_slice}
\end{table}

In this experiment, we compare our proposed method with graph cut \cite{Boykov2006}, graph cut with shape constraints \cite{Pipe_ICIP}, and DEXTR\cite{DEXTR}. For our methods and original graph cut, we use the boundary of dilation method on segmentation slice ($Seg_i$) as background seed and apply skeletonization on segmentation slice ($Seg_i$) for foreground seed. Then apply those seed point to upper slice ($i-1$) and below slice ($i+1$). This procedure also applies to DEXTR but we extract only top, bottom, leftmost, and rightmost from segmentation slice ($Seg_i$) and apply to neighbor slices. On another hand, graph cut with shape constraints method\cite{Pipe_ICIP} provide their own seed point generator for Semi-automatic 3D segmentation.

Table \ref{table:slice_by_slice} show the comparison between semi-automatic slice by slice segmentation methods. The result of DEXTR is lowest in the table. The main reason is DEXTR require accurate seed points, while the semi-automatic slice by slice segmentation will auto-generate seed points from segmentation result of the previous slice. So the result of DEXTR is over-segmentation.

On the other methods, graph cut, and graph cut with shape constraints are not required accurate seed point so the result of both methods are better than DEXTR in slice by slice segmentation. Graph cut and graph cut with shape can obtain high results on DICE, SEN, and PPV which are 0.890,0.932, and 0.855 for graph cut and 0.926, 0.908, and 0.937 for graph cut with shape constraints.

Our RefineNet can obtain high result when compared to other methods. Compare to other methods, results of the proposed method in slice by slice in Table \ref{table:slice_by_slice} and 2D slice segmentation in Table \ref{table:resnet_state} not much difference.

\begin{figure}[!tb]
  \centering
  \includegraphics[width=3.2in]{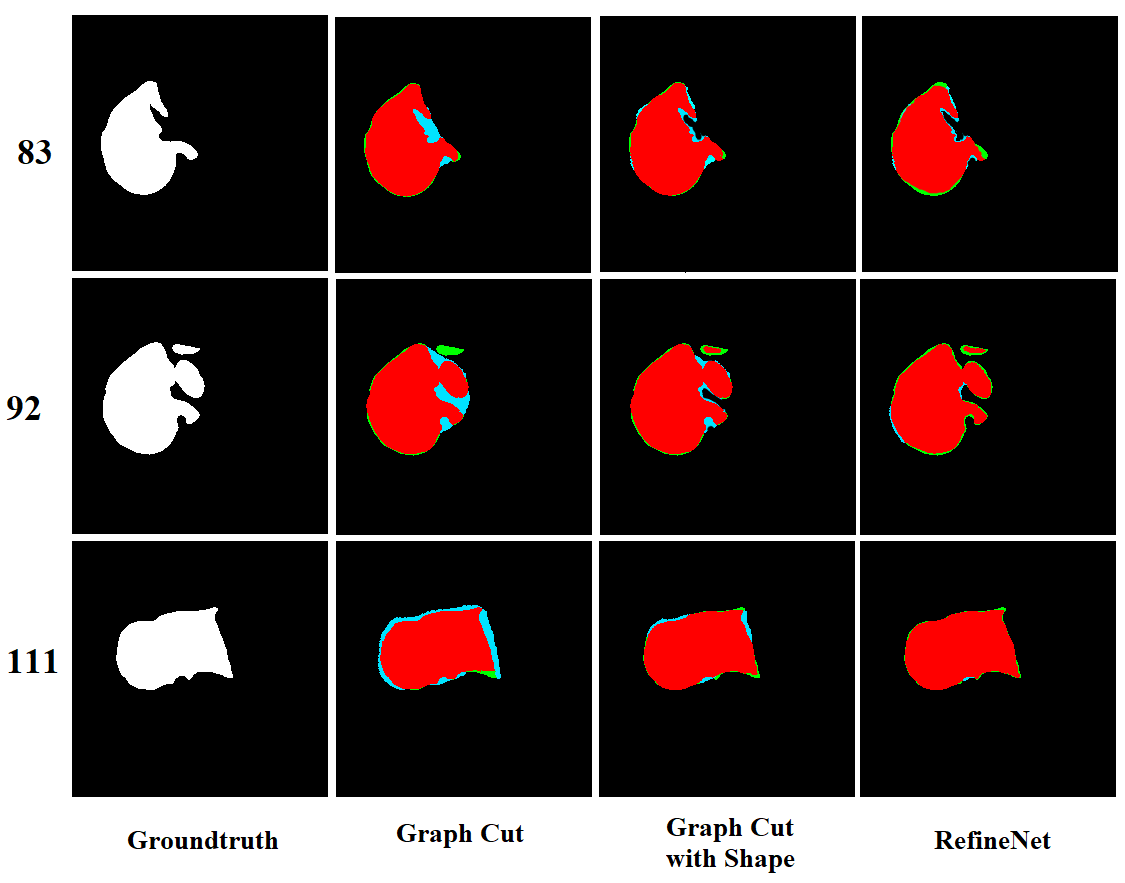}
  \caption {Visualization comparison of the slice by slice segmentation methods.}
  \label{Fig:visualization}
\end{figure}

The visualization of the slice by slice segmentation results are shown in Figure \ref{Fig:visualization}. Graph cut and graph cut with shape constraints are over-segmentation (blue color) on the slices that have a complex liver shape (slice no. 83 and 92). While graph cut shape have less under segmentation (green) than graph cut method. All methods have a high accurate segmented region (red) but our method has less under- and over-segmentation compare to both graph cut methods.

\section{Conclusion}
\label{sec:conclusion}

We proposed a RefineNet, an interactive deep refinement network for medical image segmentation. The main purpose is to refine segmentation results generated from automatic segmentation networks. The network consists of two parts. The first part is the segmentation backbone which generates the initial segmentation. While the second part is the refinement network, which combines the feature from multi-scale initial segmentation, and seed points from the user. The network can generate seed points by itself during the training phase and requires the seed points from the user only in the test phase. Besides, the proposed method can achieve high segmentation accuracy of dice coefficient in both 2D slice segmentation and semi-automatic slice by slice segmentation. Experimental results revealed that the proposed method outperforms state-of-the-art methods.


\begin{thebibliography}{8}


\bibitem{Adams1994} R. Adams, L. Bischof, Seeded Region Growing, IEEE Trans. on Pattern Analysis and Machine Intelligence, 16(6)(1994) 641--647.
\bibitem{Pan2004} Z.G. Pan, J.F. Lu, A Bayes-Based Region-Growing Algorithm for Medical Image Segmentation. Journal of Computing in Science and Engineering,9(4)(2004) 32--38.
\bibitem{Zhou2016} S. Zhou, J.Wang, S. Zhang, Y. Liang, Y. Gong, Active contour model based on local and global intensity information for medical image segmentation, Neurocomputing 186 (C) (2016) 107?118.
\bibitem{Foruzan2013} A.H. Foruzan, Y.W. Chen, R.A. Zoroofi, A. Furukawa, Y. Sato, M.Hori, N. Momiyama, Segmentation of Liver in Low-contrast Images Using K-Means Clustering and Geodesic Active Contour Algorithms, IEICE Trans. on Information and Systems, E96-D (2013) 798--807.
\bibitem{Boykov2006} Y. Boykov, G. Funka-Lea, Graph cuts and efficient n-d image segmentation, International Journal of Computer Vision 70 (2) (2006) 109?131.
\bibitem{Grady2005} L. Grady, T. Schiwietz, S. Aharon, and R. Westermann, "Randomwalks for interactive organ segmentation in two and three dimensions:implementation and validation,?in MICCAI, 2005, pp. 773?780.
\bibitem{Yuan2018} Y. Yuan, Y.-W. Chen, et al. Hybrid Method Combining Superpixel, Random Walk and Active Contour Model for Fast and Accurate Liver Segmentation. Computerized Medical Imaging and Graphics, Vol.70, pp. 119-134, 2018.
\bibitem{Okada2008} T. Okada, R. Shimada, M. Hori, M. Nakamoto, Y.-W. Chen, H. Nakamaura, Y. Sato, Automated segmentation of the liver from 3D CT images using probabilistic atlas and multi-level statistical shape model, Journal of Academic Radiology, 63(2008) 1390--1403.
\bibitem{Dong2015} C. Dong, Y.-W. Chen et al. Segmentation of liver and spleen based on computational anatomy models. Computers in biology and medicine 67: 146--160, 2015.
\bibitem{Long2015} Long, Jonathan, Evan Shelhamer, and Trevor Darrell. Fully convolutional networks for semantic segmentation. CVPR. 2015.
\bibitem{Ronneberger2015} Ronneberger, Olaf, Philipp Fischer, and Thomas Brox. U-net: Convolutional networks for biomedical image segmentation. MICCAI. 2015.
\bibitem{Abraham2018} Abraham, Nabila, and Naimul Mefraz Khan. A Novel Focal Tversky loss function with improved Attention U-Net for lesion segmentation. arXiv preprint arXiv:1810.07842, 2018.
\bibitem{Chen2014} Chen, Liang-Chieh, et al. Semantic image segmentation with deep convolutional nets and fully connected crfs. arXiv preprint arXiv:1412.7062, 2014.
\bibitem{Chen2018} Chen, Liang-Chieh, et al. Deeplab: Semantic image segmentation with deep convolutional nets, atrous convolution, and fully connected crfs. IEEE trans. on pattern analysis and machine intelligence 40.4: 834--848, 2018.
\bibitem{Chen2017} Chen, Liang-Chieh, et al. Rethinking atrous convolution for semantic image segmentation. arXiv preprint arXiv: 1706.05587, 2017.
\bibitem{Chen20182} Chen, Liang-Chieh, et al. Encoder-decoder with atrous separable convolution for semantic image segmentation. ECCV. 2018.
\bibitem{Roth2017} Roth, Holger R., et al. Hierarchical 3D fully convolutional networks for multi-organ segmentation. arXiv preprint arXiv:1704.06382, 2017.
\bibitem{He2019} He, Kelei, et al. Pelvic organ segmentation using distinctive curve guided fully convolutional networks. IEEE trans. on medical imaging 38.2: 585--595, 2019.
\bibitem{Liu2018} Liu, Xiaoming, et al. Automatic Organ Segmentation for CT Scans Based on Super-Pixel and Convolutional Neural Networks. Journal of digital imaging 31.5: 748--760, 2018.
\bibitem{Larsson2018} Larsson, Måns, Yuhang Zhang, and Fredrik Kahl. Robust abdominal organ segmentation using regional convolutional neural networks. Applied Soft Computing 70: 465--471, 2018.
\bibitem{Tang2013} Tang, Yichuan. Deep learning using linear support vector machines. arXiv preprint arXiv:1306.0239, 2013.
\bibitem{Goodfellow2014} Goodfellow, Ian, et al. Generative adversarial nets. Advances in neural information processing systems. 2014.

\bibitem{DEXTR} Maninis, K.K.,Caelles, S., Pont-Tuset, J., Van Gool, L.: Deep Extreme Cut: From Extreme Points to Object Segmentation. CVPR (2018)


\bibitem{PSP} Zhao, H., Shi, J., et al.: Pyramid scene parsing network. CVPR (2017)

\bibitem{IRCAD} 3D-IRCAD. \url{https://www.ircad.fr/research/3dircadb/}

\bibitem{UNET} Ronneberger, O., Fischer, P., Brox, T.: U-Net: Convolutional Networks for Biomedical Image Segmentation. MICCAI, 234--241 (2015)
%
%

%
\bibitem{Li2013} Li, C., Wang, X., et al.: A likelihood and local constraint level set model for liver tumor segmentation from ct volumes. IEEE Trans. Biomed. Eng. 60(10), 2967?2977 (2013)

\bibitem{Christ2016} Christ, P. F., Ettlinger, F., et al.: Automatic Liver and Tumor Segmentation of CT and MRI Volumes using Cascaded Fully Convolutional Neural Networks. MICCAI (2016)
\bibitem{Bengio2013} Bengio, Y.: Practical recommendation for gradient-based training of deep architectures. Neural Network: tricks of the trade. 437--478 (2019)

\bibitem{Pipe_ICIP} Kitrungrotsakul, T., Han, XH., Chen, YW.: Liver segmentation using superpixel-based graph cuts and restricted regions of shape constrains, IEEE International Conference on Image Processing (ICIP). 3368--3371 (2015)


\end{thebibliography}
\end{document}